\documentclass[10pt,conference]{IEEEtran}
\IEEEoverridecommandlockouts
\usepackage{cite}
\usepackage{amsmath,amssymb,amsfonts}
\usepackage{graphicx}
\usepackage{textcomp}
\usepackage{xcolor}
\usepackage{gensymb}
\usepackage{tikz}
\usepackage{caption}
\usepackage{subcaption}
\usepackage{amsmath,amssymb,graphicx}
\usepackage{xcolor}
\usepackage{algpseudocode}
\usepackage{mathtools}
\usepackage{ulem}
\usepackage{hyperref}
\hypersetup{%
    colorlinks=false,
    linkbordercolor = red,
    citebordercolor = green,
    urlbordercolor = cyan,
    pdfborder={1 1 1},
    }

\newcommand\norm[1]{\left\lVert#1\right\rVert}

\newcommand{\matr}[1]{\mathbf{#1}}

\newcommand{\mat}[1]{\boldsymbol{#1}}
\DeclareMathOperator{\Tr}{Tr}
\DeclareMathOperator{\rank}{rank}

\def\BibTeX{{\rm B\kern-.05em{\sc i\kern-.025em b}\kern-.08em
    T\kern-.1667em\lower.7ex\hbox{E}\kern-.125emX}}
\begin{document}

\title{Graph-Based Matrix Completion Applied to Weather Data}

\author{\IEEEauthorblockN{Benoît Loucheur}
\IEEEauthorblockA{\textit{ICTEAM Institute} \\
\textit{UCLouvain}\\
Louvain-la-Neuve, Belgium \\
benoit.loucheur@uclouvain.be}
\and
\IEEEauthorblockN{P.-A. Absil}
\IEEEauthorblockA{\textit{ICTEAM Institute} \\
\textit{UCLouvain}\\
Louvain-la-Neuve, Belgium \\
pa.absil@uclouvain.be}
\and
\IEEEauthorblockN{Michel Journée}
\IEEEauthorblockA{\textit{Dept. Climatology} \\
\textit{Royal Meteo. Inst. of Belgium}\\
Brussels, Belgium \\
michel.journee@meteo.be}
}

\maketitle

\begin{abstract}
Low-rank matrix completion is the task of recovering unknown entries of a matrix by assuming that the true matrix admits a good low-rank approximation. Sometimes additional information about the variables is known, and incorporating this information into a matrix completion model can lead to a better completion quality. We consider the situation where information between the column/row entities of the matrix is available as a weighted graph. In this framework, we address the problem of completing missing entries in air temperature data recorded by weather stations. We construct test sets by holding back data at locations that mimic real-life gaps in weather data. On such test sets, we show that adequate spatial and temporal graphs can significantly improve the accuracy of the completion obtained by graph-regularized low-rank matrix completion methods.
\end{abstract}


\section{Introduction}
\label{sec:intro}
Matrix completion seeks to find a low-rank matrix when only a fraction of its entries are available. This branch of research has application in numerous research projects such as collaborative filtering (Netflix challenge \cite{netflix_prize}) \cite{collaborative}, traffic sensing \cite{traf_sensing1,trafic_sensing2}, image inpainting \cite{image_inpainting}, and gene expression imputation \cite{gene_imputation}.

Sometimes it is known that there is a particular underlying structure that connects the data. Take for example the most famous case of collaborative filtering, the Netflix problem. %In this case, each row of the sparse matrix represents all movies and each column represents all users. 
In this case, each row corresponds to a user and each column to a movie.
It has been shown that this rating matrix \cite{netflix_low_rank} admits good low-rank approximations. The low-rank structure can be understood from the fact that movies can be organized in a few \textit{genres}, which are rated similarly by the same kinds of users. However, this low rank property is not the only form of structure; there can also be a graphical structure that represents connections between the row/column entities. In such a representation, the rows, resp.\ columns, are modeled as the nodes of a weighted undirected graph where the weights represent the similarity between the nodes. For example, in the case of the Netflix problem, one can create a graph that connects movies with a weight equal to the number of common actors between those movies.

%{\color{red}The regularization allows to require some additional structure in the problem variable to prevent overfitting of the model. The simplest regularization uses a Frobenius norm which ensures that the entries of the variables are penalized with a uniform weight.}

Missing data in weather time series can appear for various reasons, usually related to sensors deficiencies or communication intermittency. It was recently shown in~\cite{Loucheur} that matrix completion methods are able to outperform the state of the art for the completion of missing data in daily extreme temperature series from a network of weather stations. 

In this paper, we are interested in the application of graph-regularized matrix completion on air temperature data recorded with a 10-minute time resolution provided by the Royal Meteorological Institute (RMI) of Belgium. We show that adding a regularization by spatial and temporal graphs considerably improves the accuracy of low-rank matrix completion. We also investigate how the pattern of missing data influences the outcomes. A comparison of performance with the state of the art is also performed.

%The graph regularization admits the advantage of being able to impose non-uniform weights for the entries in the variables. With this type of data, the first graph will represent the spatial component where each node will be a weather station and the second will represent the temporal component where each node will be a timestamp of the measurement. We will study the impact of these graphs on the quality of the completion and we will show that the pattern of missing data has an influence on the results.} {\color{blue}a réécrire}

\textit{Notation:} Lowercase and uppercase boldface letters stand for vectors and matrices respectively. $\norm{\matr{A}}_F$ is the Frobenius norm of the matrix $\matr{A}$. $\norm{\matr{x}}_*$ denotes the $\ell_1$ norms of vector $\matr{x}$. Finally, $\matr{I}$ is the identity matrix and $\matr{0}$ is the zero matrix.
\section{Background and related works}
\label{sec:back}

Matrix completion is the task of recovering all the entries of a matrix $\matr{M}$ by observing only a subset $\Omega$ of them. Formally, given a matrix $\matr{M}$ of size $m\times n$, we have only access to $|\Omega| \ll m\cdot n$ entries and the goal is to predict the remaining unobserved ones in $\bar{\Omega}$. 

\subsection{Convex Models}
A classical version of the matrix completion problem consists in finding a minimum rank matrix $\matr{X}$ which is equal to the observation matrix $\matr{M}$ in the set $\Omega$:

\begin{equation}
    \begin{aligned}
        \min_{\matr{X}\in\mathbb{R}^{m\times n}} \quad & \rank(\matr{X}),\\
        \textrm{s.t.} \quad & \matr{M}_{ij}=\matr{X}_{ij}, \smallskip\forall i,j \in \Omega.
    \end{aligned}
\end{equation}
Unfortunately, solving such problem is hard \cite{np-hard}, and moreover this problem formulation is inadequate when the available data matrix is not exactly low rank due, e.g., to measurement noise. In order to reduce the complexity of this problem, \cite{np-hard} proposed to deal with the tightest possible convex relaxation of the rank operator, which is called the nuclear norm. The relaxed formulation is as follows:
\begin{equation}
    \begin{aligned}
        \min_{\matr{X}\in\mathbb{R}^{m\times n}} \quad & \norm{\matr{X}}_*\\
        \textrm{s.t.} \quad & \matr{M}_{ij} = \matr{X}_{ij}, \smallskip\forall i,j \in \Omega.
    \end{aligned}
\end{equation}
The model can then be generalized to take into account the measurement noise. The constraint is then relaxed and the optimization problem becomes:
\begin{equation}
    \begin{aligned}
        \min_{\matr{X}\in\mathbb{R}^{m\times n}} \quad & \frac{1}{2}\norm{\mathcal{P}_\Omega(\matr{M})-\mathcal{P}_\Omega(\matr{X})}_F^2 + \lambda\norm{\matr{X}}_*,
    \end{aligned}
    \label{eqn:before}
\end{equation}
where $\lambda > 0$ is a regularization parameter.

\subsection{Factorized Models}
When the sparse matrix $\matr{X}$ is large, which is usually the case for recommendation systems, it is more efficient to impose a low-rank structure on $\matr{X}$ by representing it in factorized form, $\matr{X}=\matr{A}\matr{B}^T$, where $\matr{A}$ and $\matr{B}$ have few columns.
\begin{equation}
    \begin{aligned}
        \min_{\substack{\matr{A}\in\mathbb{R}^{m\times r}\\\matr{B}\in\mathbb{R}^{n\times r}}} \quad & \frac{1}{2}\norm{\mathcal{P}_\Omega(\matr{M})-\mathcal{P}_\Omega(\matr{A}\matr{B}^T)}_F^2\\
        &+ \frac{\lambda_a}{2}\norm{\matr{A}}_F + \frac{\lambda_b}{2}\norm{\matr{B}}_F,
    \end{aligned}
    \label{eqn:after}
\end{equation}
where $\lambda_a,\lambda_b > 0$ are two regularization parameters and $r$ is the target rank of the matrix.

%To make the transition from (\ref{eqn:before}) to (\ref{eqn:after}), we must also use a useful property of the nuclear norm \cite{nuclear_norm}:
Formulation~\eqref{eqn:before} and~\eqref{eqn:after} are closely related in view of the following property of the nuclear norm \cite{nuclear_norm,SVD}:
\begin{equation}
    \begin{aligned}
    \norm{\matr{X}}_* = \min_{\substack{\matr{A}\in\mathbb{R}^{m\times r}\\\matr{B}\in\mathbb{R}^{n\times r}}} &\norm{\matr{A}}_F^2+\norm{\matr{B}}_F^2\\
    \textrm{s.t.} \quad &\matr{X} = \matr{A}\matr{B}^T.
    \end{aligned}
\end{equation}

\subsection{Graph-based Regularization}
For simplicity, we present the case of a graph on the rows of $\matr{M}$; since the columns of $\matr{M}$ are the rows of $\matr{M}^T$, the extension to a graph on the columns of $\matr{M}$ is direct. Hence assume the availability of a graph $\mathcal{G}^a = (\mat{V}^a,\mat{E}^a,\matr{W}^a)$ on the rows with vertices $\mat{V}^a=\{1,\cdots,m\}$, edges $\mat{E}^a\subseteq \mat{V}^a\times \mat{V}^a$ and non-negative weights on the edges represented by the symmetric $m \times m$ matrix $\matr{W}^a$. If there is an edge between the nodes $i$ and $j$, then $\matr{W}_{ij}^a=\matr{W}_{ji}^a\neq 0$.
This information in the form of a graph can then be incorporated into the minimization~\eqref{eqn:laplacian} using the following term \cite{graphmin}:
\begin{equation}
    \begin{aligned}
        \frac{1}{2}\sum_{i,j}\matr{W}_{ij}^{a}\norm{\matr{a}_i-\matr{a}_j}_2^2=\Tr\big(\matr{A}^T\mathbf{Lap}(\matr{W}^a)\matr{A}\big),
    \end{aligned}
    \label{eqn:laplacian}
\end{equation}
where $\mathbf{Lap}(\matr{W}^a) = \matr{D}^a - \matr{W}^a$ is the graph Laplacian of $\matr{W}^a$, $\matr{D}^a$ is the diagonal degree matrix, with $\matr{D}_{i,i}^a=\sum_{i\neq j}\matr{W}_{i,j}^a$. The left-hand side term of ($\ref{eqn:laplacian}$) favors a similarity between the rows $\matr{a}_i$ and $\matr{a}_j$ of $\matr{A}$ whenever $\matr{W}_{ij}^{a}$ is large. The reasoning is the same for the graph on the columns $\mathcal{G}^b$.

In view of the above, the problem of matrix completion over graphs can be formulated as follows \cite{graph_form}:
\begin{align}
    &\min_{\substack{\matr{A}\in\mathbb{R}^{m\times r}\\\matr{B}\in\mathbb{R}^{n\times r}}} \quad  \frac{1}{2}\norm{\mathcal{P}_\Omega(\matr{M})-\mathcal{P}_\Omega(\matr{A}\matr{B}^T)} \nonumber\\
    &+ \frac{\lambda_L}{2}\Big\{\Tr\big(\matr{A}^T\mathbf{Lap}(\matr{W}^a)\matr{A}\big) + \Tr\big(\matr{B}^T\mathbf{Lap}(\matr{W}^b)\matr{B}\big)\Big\}\nonumber\\
     &\hspace{3.8cm}+\frac{\lambda_a}{2}\norm{\matr{A}}_F^2 + \frac{\lambda_b}{2}\norm{\matr{B}}_F^2. \label{eqn:grmf}
\end{align}
This formulation having two terms of regularization by graphs, it is more commonly called Graph-Regularized Matrix Completion (GRMC).
%Unfortunately, the computational load quickly becomes prohibitive for large input matrices $\matr{M}$. Instead, the authors of \cite{GRALS} provide an efficient implementation (called GRALS) by applying the conjugate gradient method alternatively to $\matr{A}$ and $\matr{B}$.
A method called GRALS, to address~\eqref{eqn:grmf} by applying a conjugate gradient method alternatively to $\matr{A}$ and $\matr{B}$, is provided in~\cite{GRALS}.

\subsection{State of the Art}
%In order to compare the performance of graph regularization, we will compare it with the state of the art.
In~\cite{Loucheur}, the following methods were compared on weather data completion tasks.

IDW (Inverse Distance Weighting) is a widely-used method for estimating missing data in meteorological and other time series datasets. The approach uses a weighted average of neighboring observed data points, with weights computed by the inverse distance between the missing data points and the neighbors. IDW is simple,  computationally efficient, and can produce accurate estimation when the underlying data has a smooth spatial structure. However, its performance is affected by outliers or uneven data distributions.

PCA~\cite{PCA} (Principal Component Analysis) is another method for the estimation of missing data. The approach involves decomposing the data into its principal components, which capture the most important patterns or variation in the data. Then, the missing data can be estimated using a linear combination of the principal components based on the observed values in the corresponding time periods.

SoftImpute~\cite{SoftImpute} is a matrix completion algorithm that uses the Singular Value Decomposition (SVD) to estimate the low-rank structure and then applies a soft-thresholding operator to shrink small singular values. The formulation of SoftImpute is as follow:
\begin{equation}
    \begin{aligned}
        \min_{\matr{X}\in\mathbb{R}^{m\times n}} \quad & \norm{\matr{X}}_*,\\
        %\textrm{s.t.} \quad & \mathcal{P}_\Omega(\matr{M}-\matr{X}) \leq \delta.\\
        \textrm{s.t.} \quad & \mathcal{P}_\Omega(\matr{M})-\mathcal{P}_\Omega(\matr{X}) \leq \delta.
    \end{aligned}
    \label{eqn:softimpute}
\end{equation}

Finally, RTRMC~\cite{RTRMC_ABSIL_BOUMAL} addresses the matrix completion problem by applying a trust-region methods to the following optimization problem on the Grassmann manifold $\mathcal{G}^{m\times r}$:

%Finally, RTRMC \cite{RTRMC_ABSIL_BOUMAL} is a matrix completion method that assumes the complete data matrix lies on the Riemannian manifold, which is a space with a curved geometry. To optimize the cost function, RTRMC uses a trust-region approach that finds a local minimum within a trust-region around the current iterate. The formulation of RTRMC is as follow:
\begin{equation}
        \min_{\mathcal{A}\in\mathcal{G}^{m\times r}}\min_{\matr{B}\in\mathbb{R}^{n\times r}} \hspace{0.2em} \frac{1}{2}\norm{\mathcal{P}_\Omega(\matr{M})-\mathcal{P}_\Omega(\matr{A}\matr{B}^T)}_F^2 + \frac{\lambda^2}{2} \norm{\mathcal{P}_{\bar{\Omega}}(\matr{A}\matr{B}^T)}_F^2,
\end{equation}
where $\bar{\Omega}$ is the complement of the set $\Omega$.

\section{Graph Generation}
In addition to the estimation of the classical hyperparameters of a matrix completion method, we also need to generate two graphs, one for the rows of the matrix (i.e., each timestamp) and one for the columns (i.e., each station). In this subsection, different ways to create our spatial and temporal graphs are described. 
\subsection{Spatial Graph}
The graph of the matrix columns is in our case the spatial graph.  Each node represents a weather station and we are free to create a weighted edge or not between two stations. We have compared different graph possibilities with various levels of complexity. 

\begin{figure}[!h]
    \centering
    \begin{subfigure}{.45\linewidth}
        \includegraphics[width=1\textwidth]{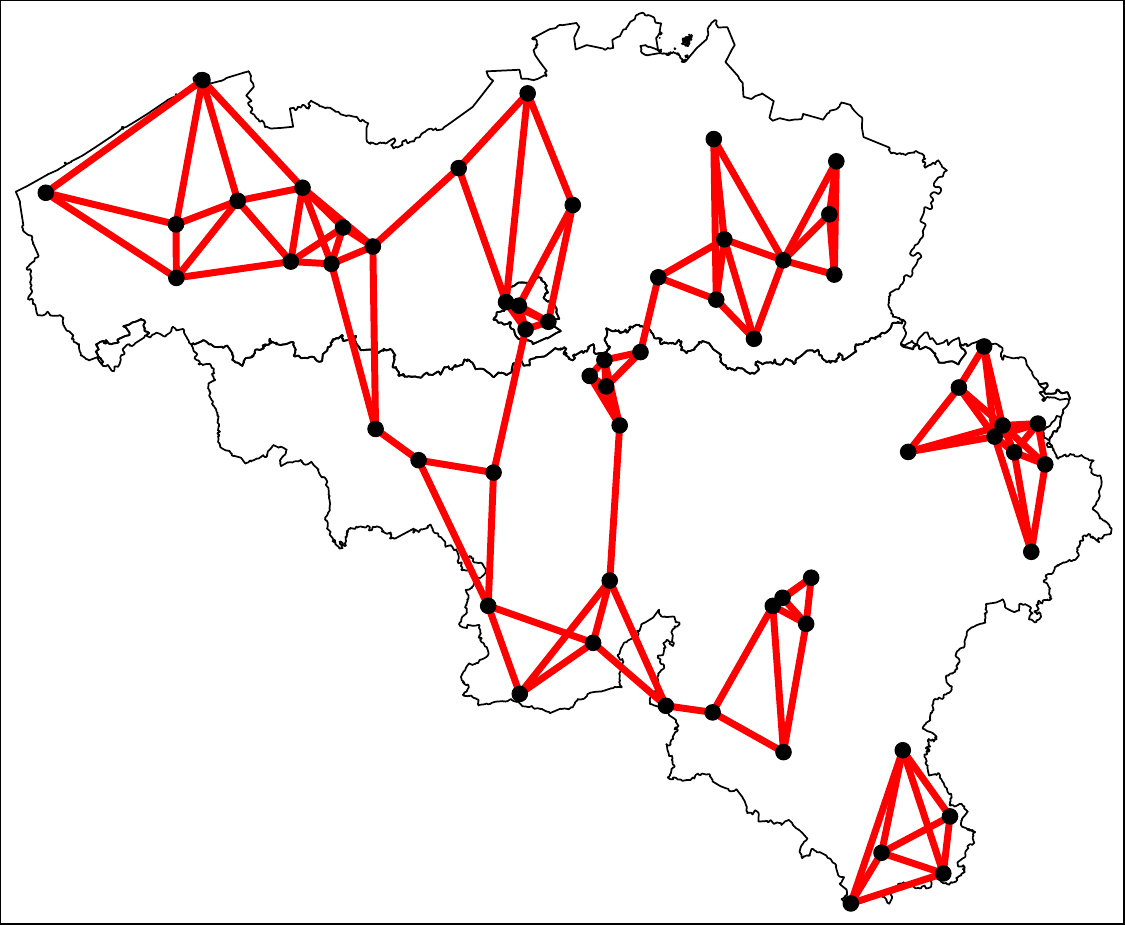}
        \caption{3-KNN}
        \label{fig:KNN}
    \end{subfigure}
    \hspace{0.5cm}
    \begin{subfigure}{.45\linewidth}
        \includegraphics[width=1\textwidth]{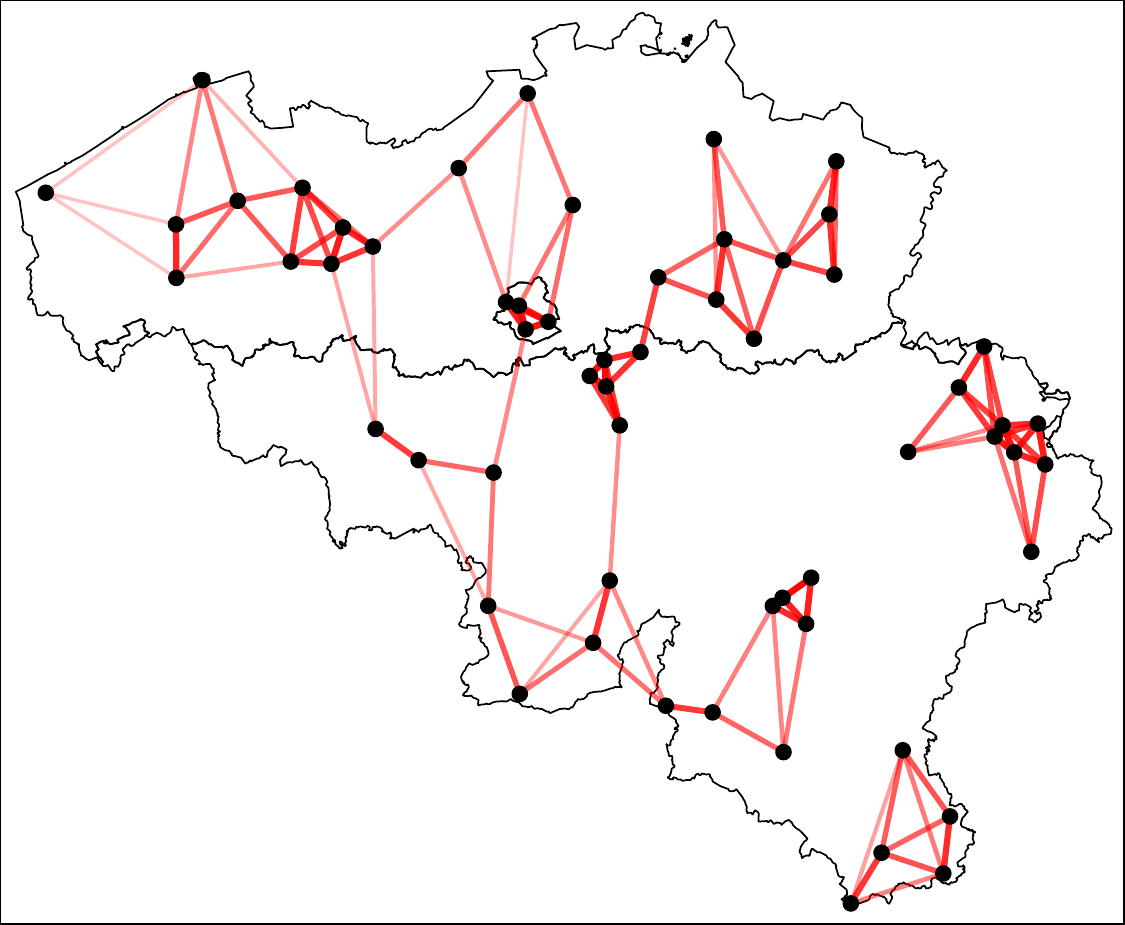}
        \caption{3-KNN weighted}
        \label{fig:KNN_WEIGHTED}
    \end{subfigure}
    \newline
    \begin{subfigure}{.49\linewidth}
        \includegraphics[width=1\textwidth]{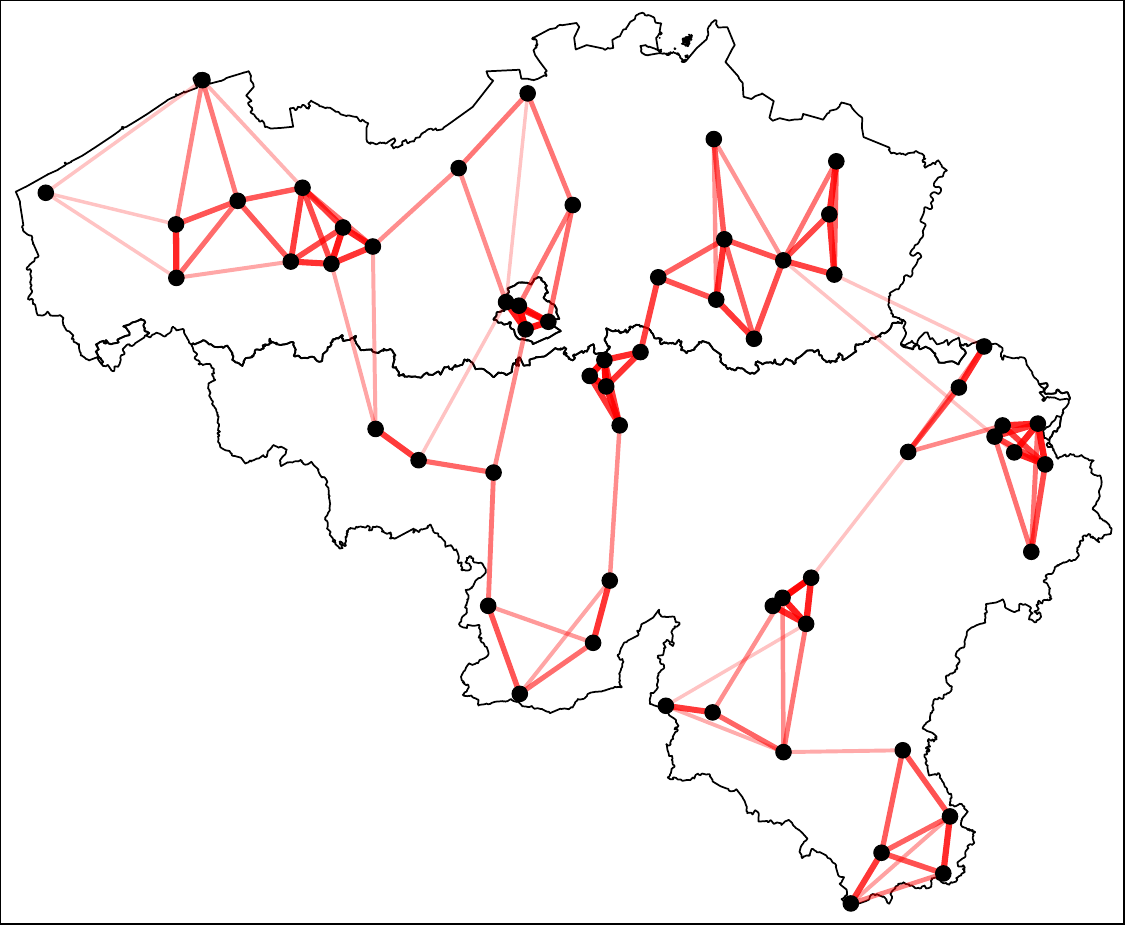}
        \caption{3-KNN weighted with altitude limit}
        \label{fig:KNN_WEIGHTED_ALT}
    \end{subfigure}
    \caption{Spatial Graph examples}
\end{figure}

The most basic approach is the K-Nearest Neighbors (KNN), in which case each station is connected by an edge to its K nearest neighbors using the geographic distance. (see Fig.~\ref{fig:KNN}). 
It is then possible to add constraints on the generation of the KNN graph by adding as weights for the edges the inverse of the distance between the two stations. Thus, the closer the stations are, the more strongly they are connected (see Fig.~\ref{fig:KNN_WEIGHTED}). The visualization of the weights in the graph is done via the transparency and thickness of the edges. The more important the weight is, the darker and thicker the edge will be.
A third version of the spatial graph consists in adding a constraint of maximum altitude difference between two stations. If the altitude difference between two stations is greater than a chosen threshold value, then this connection is ignored. This avoids the rather special case where two stations are very close spatially but have a large difference in altitude, e.g., more than 100 meters (see Fig.~\ref{fig:KNN_WEIGHTED_ALT}). Such cases can appear in areas with complex orography and can negatively impact the completion of missing data. %As it can be seen by comparing Figure \ref{fig:KNN_WEIGHTED} and \ref{fig:KNN_WEIGHTED_ALT}, the southern part of Belgium is rather hilly, which implies new connections for the edges. There is no change for the northern part which is rather made of a rather flat relief. The edges in the northern part of Belgium do not change when an altitude limit condition is imposed, due to the fairly flat terrain there.\\
By comparing the Fig.~\ref{fig:KNN_WEIGHTED} and Fig.~\ref{fig:KNN_WEIGHTED_ALT}, no changes are observed for the edges in the north of Belgium. This is due to the relatively flat terrain in this geographical area which will then never trigger the altitude limit condition. In contrast, the southern part of Belgium has a much more complex orography which causes changes in the edges. 

\subsection{Temporal Graph}
The graph of the matrix rows is in our case the temporal graph. Each node represents a date and to link them we use the notion of lag set $\mathcal{L}$ which is the (repeating) dependency pattern of the graph.

\begin{figure}[!h]
	\begin{subfigure}{.48\textwidth}
		\centering
		\begin{tikzpicture}
    \node at (0,0) [circle,draw,scale=1]{$t_1$};
    \node at (1.5,0) [circle,draw]{$t_2$};
    \node at (3,0) [circle,draw]{$t_3$};
    \node at (4.5,0) [circle,draw]{$t_4$};
    \draw [color=red!100,line width=1pt](0.4,0) to node[midway,above=-2.5pt] {$w_1$} (1.1,0);
    \draw [color=red!100,line width=1pt](1.9,-0) to node[midway,above=-2.5pt] {$w_1$} (2.6,0);
    \draw [color=red!100,line width=1pt](3.4,-0) to node[midway,above=-2.5pt] {$w_1$} (4.1,0);
    
    \node at (5.5,0){$\cdots$};
\end{tikzpicture}
		\caption{$\mathcal{L} = [1]$}
		\label{fig:TRMF1}
	\end{subfigure}
	\newline
	\newline
	\begin{subfigure}{.48\textwidth}
		\centering
		\begin{tikzpicture}
    \node at (0,0) [circle,draw,scale=1]{$t_1$};
    \node at (1.5,0) [circle,draw]{$t_2$};
    \node at (3,0) [circle,draw]{$t_3$};
    \node at (4.5,0) [circle,draw]{$t_4$};
    \draw [color=red!100,line width=1pt](0.4,0) to node[midway,above=-2.5pt] {$w_1$} (1.1,0);
    \draw [color=red!100,line width=1pt](1.9,-0) to node[midway,above=-2.5pt] {$w_1$} (2.6,0);
    \draw [color=red!100,line width=1pt](3.4,-0) to node[midway,above=-2.5pt] {$w_1$} (4.1,0);
    \draw [color=blue!100,line width=1pt](0.2828,-0.2828)  [bend right=45] to node[midway,above=-2.5pt] {$w_2$} (3-0.2828,-0.2828);
    \draw [color=blue!100,line width=1pt](1.5+0.2828,-0.2828)  [bend right=45] to node[midway,above=-2.5pt] {$w_2$} (4.5-0.2828,-0.2828);
    \draw [color=black!60!green,line width=1pt](0,-0.4) [bend right=45] to node[midway,above=-2.5pt]{$w_3$} (4.5,-0.4);
    
    \node at (5.5,0){$\cdots$};
\end{tikzpicture}
		\caption{$\mathcal{L} = [1,2,3]$}
		\label{fig:TRMF2}
	\end{subfigure}
	\caption{Temporal Graph with different lag sets $\mathcal{L}$}
	\label{fig:TRMF}
\end{figure}
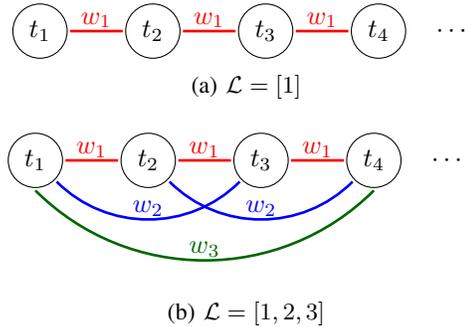

Fig. \ref{fig:TRMF1} shows the simplest lag set when $\mathcal{L}=[1]$. This means that the temperature measured at time t will be connected to the temperature at time $t-10$ minutes and $t+10$ minutes.

Fig. \ref{fig:TRMF2} shows the lag set $\mathcal{L}=[1,2,3]$, which creates a link between the current measurement and the three previous and following ones.

The first lag set only tries to capture the immediate relationship or dependency between two consecutive measures, whereas the second lag set tries to capture a slightly longer-term relationship.

Once the temporal graph is created, we have the possibility to choose weights for its edges. In our experiments, we chose two different rules for the attribution of weights. The first trivial one is when $w_i = 1$ which allows us to have in a non-weighted graph. The other possibility considered is to use the rule $w_i=\frac{1}{i}$ to emphasize the importance of short-term relationships in the data, as shorter lags often reflect more immediate or direct dependencies, while longer lags may be more influenced by noise or other indirect factors.

\section{Machine Learning Model}
All our experiments were performed with data provided by the RMI, consisting of air temperature measurements from 50 automatic weather stations located in Belgium, which provide data with a temporal resolution of 10-min. We consider in this study the data for the period from January 1, 2020 to March 1, 2022. The source code is available at \url{https://github.com/bloucheur/GRMCWeatherData}. Implementations of the existing algorithms are also publicly available, however the dataset we use is confidential. An open source dataset containing similar data is available.

\subsection{Generation of Missing Weather Data}
\begin{figure}[h!]
    \centering
    \begin{subfigure}{.37\linewidth}
        \includegraphics[width=1\textwidth]{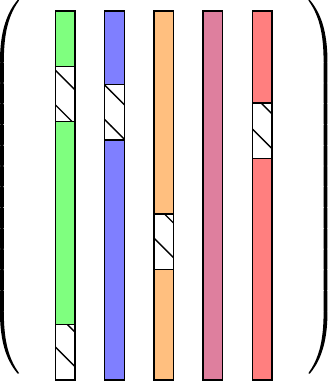}
        \caption{Bloc{k}}
        \label{fig:bloc_di}
    \end{subfigure}
    \hspace{0.5cm}
    \begin{subfigure}{.37\linewidth}
        \includegraphics[width=1\textwidth]{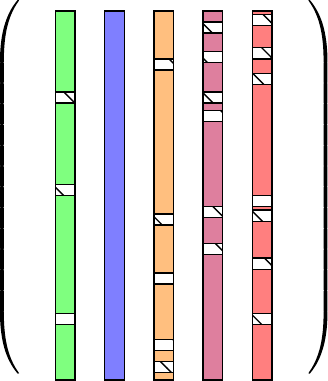}
        \caption{Spread}
        \label{fig:spread_di}
    \end{subfigure}
    \caption{The two scenarios considered, the hatched areas represent the missing data.}
    \label{fig:scenario}
\end{figure}

To evaluate the performance of our model, we have to synthetically create missing entries in the matrix $\matr{M}$. These synthetic missing data will be used to evaluate the error committed by our model once the completion of these missing data is calculated. However, as we will show in our results in Section \ref{ssec:results}, the pattern of missing data significantly impacts the quality of the completion. In our case, two types of patterns are considered. The first one, represented in Fig.~\ref{fig:bloc_di}, considers that the missing data are in the form of long duration blocks (between one and 3 days). While the second scenario, represented in Fig.~\ref{fig:spread_di}, considers synthetic holes of small duration (10 minutes to 2 hours). It is important to note that in total the number of missing data generated for both scenarios is the same. Furthermore, the number of artificially generated missing data is 10\% of the maximum number of entries in the matrix, more formally: $|\Omega_{\text{Block}}| = |\Omega_{\text{Spread}}|= 0.1*mn$.

Note that this condition is not applied in Fig.~\ref{fig:bloc_di} and Fig.~\ref{fig:spread_di} as these represent an artist's view of our two scenarios.

\subsection{Hyperparameter Tuning}
The graph regularized matrix completion model we study, named GRALS\cite{GRALS}, has hyperparameters to choose from. In particular, the value of the rank ($r$) for the factorization, the three constants of regularization ($\lambda_L,\lambda_a$ and $\lambda_b$) as well as variables of our design for the generation of our graphs which are explained above. We also want to simulate our missing data according to particular patterns as shown in Fig.~\ref{fig:scenario}. The machine learning model that corresponds to our expectations is the Monte Carlo Cross-Validation~\cite{montecarlo}. The advantage of this evaluation technique is that it allows us to simulate missing data according to a pattern that we can define.

We first split our datasets in two, the data from January 1, 2020 to August 31, 2021 are part of the training set, the rest (i.e. September 1, 2021 to March 1, 2022) are part of the testing set. 
%{\color{red}For each of our experiments, we select one week of data for all our stations. We thus have a matrix $\matr{M}\in\mathbb{R}^{1009\times 50}$ containing no missing values, thus $\bar{\Omega}=\{\emptyset\}$.}
For each experiment, we consider one week of data for all stations, i.e., the observation matrix $\matr{M}\in\mathbb{R}^{1009\times 50}$.
In order to perform the hyperparameter tuning, we select $10$ weeks without gaps and that do not intersect in the training set. For each of these $10$ weeks, we generate $5$ patterns of missing data, to create our sparse matrix $\matr{M}$, according to the scenario imposed for the experiment. The optimal set of hyperparameters is determined as the one that results into the lowest RMSE when averaged over the $5\times 10$ folds. It is important to note that the weeks selected for the two scenarios are the same, only the way the missing data are generated differs.
For the final evaluation on the test set, the same missing data generation methodology is applied with $5$ different weeks and $3$ generations of missing data patterns.
\vfill\null
%\columnbreak
\section{Experimental Results}
\label{ssec:results}

\subsection{Comparison of the methods}
TABLE \ref{tab:compa} shows the values of RMSE for every method on the test set.
\begin{table}[h!]
    \centering
    \begin{tabular}{|c||c|c|}
        \hline
         Methods & Block & Spread \\
         \hline
         IDW& 0.61 & 0.57\\
         \hline
         PCA& 0.78&0.71\\
         \hline
         SoftImpute & 0.50 & 0.41 \\
         \hline
         RTRMC & 0.48 & 0.50 \\
         \hline
         GRALS & 0.51 & 0.45 \\
         \hline
    \end{tabular}
    \caption{Average RMSE (in °C) on the test set for every methods considered. Results
for the two types of missing data generation with different
constraints applied on the model.}
\label{tab:compa}
\end{table}

The results indicate that the performance of GRALS, under our experimental conditions, is comparable to the other low-rank matrix completion methods (SoftImpute and RTRMC). 

In the remainder of this experimental section, we will conduct ablation studies on GRALS to evaluate how much the achieved RMSE depends on its spatial and temporal graph regularization.
%In the remainder of this experimental section, we will analyse in detail the performance of GRALS to evaluate whether the addition of the graph improves the performance of a classical matrix completion model.

\subsection{Graph Part}
TABLE~\ref{tab:hyp} represents all the hyperparameters of the graph regularized matrix completion model that we have. For each hyperparameter, we define a set of possible and relevant values. Due to the high number of hyperparameters in our model, we generate all possible combinations of hyperparameters, and a sampling without replacement is performed. This approach is based on the RandomizedSearchCV model from Scikit-learn~\cite{scikit}.

\begin{table}[!h]
    \centering
    \begin{tabular}{|l|l||l|l|}
        \hline
        Hp&Values tested&Block&Spread\\
        \hline
        $r$&$\{2, \cdots,20\}$&10&11\\
        \hline
        $\lambda_L$&$\{0.1,0.01,0.001,0.005\}$&0.001&0.001\\
        \hline
        $\lambda_a$&$\{0.1,0.01,0.001,0.005\}$&0.005&0.001\\
        \hline
        $\lambda_b$&$\{0.1,0.01,0.001,0.005\}$&0.005&0.005\\
        \hline
        $K$&$\{1,2, \cdots,5\}$&4&3\\
        \hline
        weighted&$\{\text{True},\text{False}\}$&True&True\\
        \hline
        altitude\_limit&$\{\text{True},\text{False}\}$&True&True\\
        \hline
        $\mathcal{L}$&$\{[1],[1,2],[1,2,3],[1,2,3,4,5]\}$&[1]&[1,2,3]\\
        \hline
    \end{tabular}
    \caption{Set of hyperparameters (Hp) with their optimal values for the Block and Spread missing data patterns on the training set.}
    \label{tab:hyp}
\end{table}
The last two columns of this table represent the optimal hyperparameters obtained during the training phase for the two scenarios considered.

Whether it is the block or spread scenario, the spatial graph requires a large number of $K$ neighbours and it is important to note that both require a weighted spatial graph coupled with the altitude boundary condition.

TABLE~\ref{tab:test_results} represents different RMSE values on the test set by imposing or not a priori conditions on some hyperparameters. The Case \#1 does not impose any constraints, and we consider these results as a reference.

\begin{table}[!h]
    \centering
    \begin{tabular}{|c|c|c|c|}
        \cline{2-4}
        \multicolumn{1}{l|}{}&GRALS Conditions&Block&Spread\\
        \hline
         Case \#1&No constraints&0.51&0.45  \\
         \hline
         Case \#2&$\lambda_L=\lambda_a=\lambda_b=0$ & 0.95&0.93\\
         \hline
         Case \#3&$\lambda_a=\lambda_b=0$&0.90&0.91\\
         \hline
         Case \#4&$\lambda_L=0$&0.76&0.67\\
         \hline
         Case \#5&$\matr{Lap}(\matr{W}^b)=\matr{0}$&0.71&0.61\\
         \hline
         Case \#6&$\matr{Lap}(\matr{W}^a)=\matr{0}$&0.55&0.6\\
         \hline
    \end{tabular}
    \caption{Average RMSE (in \degree C) on the test set. Results for the two types of missing data generation with different constraints applied on the model.}
    \label{tab:test_results}
\end{table}

Setting all regularization terms to zero ($\lambda_L=\lambda_a=\lambda_b=0$) drastically decreases the performance of the completion regardless of the missing data pattern scenario.

%One of the most interesting parts of the results is the interpretation of the last two rows of the table.  What we observe is that cancelling the contribution of the temporal graph (i.e., by imposing $\matr{Lap}(\matr{W}^a)=\matr{I}$) has very small impact on the results when the scenario is Bloc. Whereas in the case where the missing data scenario is Spread, cancelling this graph increases the RMSE from 0.45 to 0.60. 

%When we cancel the presence of the spatial graph (i.e., by imposing $\matr{Lap}(\matr{W}^b)=\matr{I}$), the error increases significantly in both types of scenarios. This is a rather expected result, the layout of the weather stations as well as the continuous nature of the temperature in space, conforms us in the idea that linking stations close in space could improve the results.
Case \#5 and Case \#6 show the impact of considering only a temporal graph ($\matr{Lap}(\matr{W}^b)=\matr{0}$) or only a spatial graph ($\matr{Lap}(\matr{W}^a)=\matr{0}$). A comparison between Case \#1 and Case \#6 shows that the temporal graph contributes little to the resulting completion performance in the case of the Block scenario. On the other hand, in the Spread scenario, removing the temporal graph significantly impacts the completion with an average RMSE that increases by $35\%$. 

Then, by comparing Case \#1 and Case \#5, we see that the spatial graph is important for both scenarios. This is a rather expected result: the layout of the weather stations as well as the continuous nature of the temperature in space, reinforces the idea that linking stations close in space could improve the results.
%For the comparison between the two scenarios, the most interesting result is to analyze the impact of the absence of the spatial and/or temporal graph. When considering the block scenario, removing the temporal graph (i.e., impose \textbf{$\matr{Lap}(\matr{E}^u)=\matr{I}$}) does not impact the results much. This graph does not have a huge impact on the completion. Whereas in the spread scenario, removing the temporal graph has much more impact. This result is due to  the way the temporal graph is constructed.

\normalem

\section{Conclusion}
It is known~\cite{Loucheur} that the low-rank matrix completion methods SoftImpute and RTRMC are very effective for completing missing meteorological data. Our experiments (Section~\ref{ssec:results}) have shown that a \emph{graph-regularized} low-rank matrix completion method, termed GRALS, is also very effective, and that its graph regularization is essential to obtain competitive results. This opens two avenues of research: (i) enhance RTRMC and SoftImpute with graph regularization and (ii) improve the construction of the graphs.

Currently, our graphs are built using only the metadata at our disposal (i.e., geographic coordinates of the stations as well as their altitudes). In our future research, we will attempt to build adaptive graphs based on the measurements. By defining the notion of distance between two time series, it is possible to perform clustering and thus generate a spatial graph. For the temporal graph, it is possible to fit an autoregressive model on the known data, thus giving us weight values for each weight of the lag set.
\label{sec:conc}

\vfill\null
%\columnbreak
% \begin{table}[h!]
%     \centering
%     \begin{tabular}{|c||c|c|}
%         \hline
%          Methods & Block & Spread \\
%          \hline
%          PCA& 0.78&0.71\\
%          \hline
%          IDW& 0.61 & 0.57\\
%          \hline
%          RTRMC & 0.48 & 0.50 \\
%          \hline
%          SoftImpute & 0.50 & 0.41 \\
%          \hline
%          GRALS & 0.51 & 0.45 \\
%          \hline
%     \end{tabular}
%     \caption{Caption}
%     \label{tab:my_label}
% \end{table}

%\section{REFERENCES}
%\label{sec:refs}

%\vfill
%\pagebreak


\bibliographystyle{IEEEbib}
\bibliography{bibli.bib}

\begin{thebibliography}{10}

\bibitem{netflix_prize}
James Bennett, Stan Lanning, and Netflix,
\newblock ``The {Netflix} prize,''
\newblock in {\em In KDD Cup and Workshop in conjunction with KDD}, 2007.

\bibitem{collaborative}
Jasson D.~M. Rennie and Nathan Srebro,
\newblock ``Fast maximum margin matrix factorization for collaborative
  prediction,''
\newblock in {\em Proceedings of the 22nd International Conference on Machine
  Learning}, New York, NY, USA, 2005, ICML '05, p. 713–719, Association for
  Computing Machinery.

\bibitem{traf_sensing1}
Rong Du, Cailian Chen, Bo~Yang, Ning Lu, Xinping Guan, and Xuemin Shen,
\newblock ``Effective urban traffic monitoring by vehicular sensor networks,''
\newblock {\em IEEE Transactions on Vehicular Technology}, vol. 64, no. 1, pp.
  273--286, 2015.

\bibitem{trafic_sensing2}
Rong Du, Cailian Chen, Bo~Yang, and Xinping Guan,
\newblock ``Vanet based traffic estimation: A matrix completion approach,''
\newblock in {\em 2013 IEEE Global Communications Conference (GLOBECOM)}, 2013,
  pp. 30--35.

\bibitem{image_inpainting}
Hongyang Xue, Shengming Zhang, and Deng Cai,
\newblock ``Depth image inpainting: Improving low rank matrix completion with
  low gradient regularization,''
\newblock {\em {IEEE} Transactions on Image Processing}, vol. 26, no. 9, pp.
  4311--4320, sep 2017.

\bibitem{gene_imputation}
Arnav Kapur, Kshitij Marwah, and Gil Alterovitz,
\newblock ``Gene expression prediction using low-rank matrix completion,''
\newblock {\em BMC Bioinformatics}, vol. 17, 06 2016.

\bibitem{netflix_low_rank}
Robert Bell and Yehuda Koren,
\newblock ``Lessons from the {Netflix} prize challenge,''
\newblock {\em SIGKDD Explorations}, vol. 9, pp. 75--79, 12 2007.

\bibitem{Loucheur}
Benoît Loucheur, P.-A. Absil, and Michel Journée,
\newblock ``Gap filling in air temperature series by matrix completion
  methods,''
\newblock in {\em Proceedings of the 30th European Symposium on Artifical
  Neural Networks, Computational Intelligence and Machine Learning
  (ESANN2022)}. 2022, pp. 369--374, i6doc.

\bibitem{np-hard}
Emmanuel~J. Cand\`{e}s and Benjamin Recht,
\newblock ``Exact matrix completion via convex optimization,''
\newblock {\em Found. Comput. Math.}, vol. 9, no. 6, pp. 717–772, dec 2009.

\bibitem{nuclear_norm}
Benjamin Recht, Maryam Fazel, and Pablo~A. Parrilo,
\newblock ``Guaranteed minimum-rank solutions of linear matrix equations via
  nuclear norm minimization,''
\newblock {\em {SIAM} Review}, vol. 52, no. 3, pp. 471--501, jan 2010.

\bibitem{SVD}
Trevor Hastie, Rahul Mazumder, Jason~D. Lee, and Reza Zadeh,
\newblock ``Matrix completion and low-rank {SVD} via fast alternating least
  squares,''
\newblock {\em Journal of Machine Learning Research}, vol. 16, no. 104, pp.
  3367--3402, 2015.

\bibitem{graphmin}
Mikhail Belkin and Partha Niyogi,
\newblock ``Laplacian eigenmaps for dimensionality reduction and data
  representation,''
\newblock {\em Neural Computation}, vol. 15, no. 6, pp. 1373--1396, 2003.

\bibitem{graph_form}
Vassilis Kalofolias, Xavier Bresson, Michael~M. Bronstein, and Pierre
  Vandergheynst,
\newblock ``Matrix completion on graphs,''
\newblock {\em CoRR}, vol. abs/1408.1717, 2014.

\bibitem{GRALS}
Nikhil Rao, Hsiang-Fu Yu, Pradeep~K. Ravikumar, and Inderjit~S. Dhillon,
\newblock ``Collaborative filtering with graph information: Consistency and
  scalable methods,''
\newblock in {\em Advances in Neural Information Processing Systems 27}, 2015.

\bibitem{PCA}
Hege Hisdal and Ole Tveito,
\newblock ``Extension of runoff series using empirical orthogonal functions,''
\newblock {\em Hydrological Sciences Journal}, vol. 38, pp. 33--49, 02 1993.

\bibitem{SoftImpute}
Jian-Feng Cai, Emmanuel~J. Cand\`{e}s, and Zuowei Shen,
\newblock ``A singular value thresholding algorithm for matrix completion,''
\newblock {\em SIAM Journal on Optimization}, vol. 20, no. 4, pp. 1956--1982,
  2010.

\bibitem{RTRMC_ABSIL_BOUMAL}
N.~Boumal and P.-A. Absil,
\newblock ``{RTRMC: A Riemannian trust-region method for low-rank matrix
  completion},''
\newblock in {\em Advances in Neural Information Processing Systems 24
  ({NIPS})}, pp. 406--414. 2011.

\bibitem{montecarlo}
Richard~R. Picard and R.~Dennis Cook,
\newblock ``Cross-validation of regression models,''
\newblock {\em Journal of the American Statistical Association}, vol. 79, no.
  387, pp. 575--583, 1984.

\bibitem{scikit}
James Bergstra and Yoshua Bengio,
\newblock ``Random search for hyper-parameter optimization,''
\newblock {\em J. Mach. Learn. Res.}, vol. 13, pp. 281–305, feb 2012.

\end{thebibliography}

\end{document}